\documentclass[runningheads]{llncs}

\usepackage[T1]{fontenc}
\IfFileExists{glyphtounicode.tex}{\input{glyphtounicode}\pdfgentounicode=1}{}
\usepackage{amsmath,amssymb}
\usepackage{booktabs}
\usepackage{graphicx}
\usepackage{placeins}
\usepackage{xcolor}
\usepackage{url}
\usepackage{orcidlink}
\hypersetup{hidelinks}

\graphicspath{{figures/}}

\newcommand{\latent}{\mathbf{z}}
\newcommand{\coord}{\mathbf{x}}

\newcommand{\context}{\mathbf{c}}
\newcommand{\obs}{\mathcal{O}}
\newcommand{\stdlatent}[1]{\tilde{\latent}(#1)}

\newcommand{\loss}{\mathcal{L}}

\title{Observation-Conditioned Latent Energy Priors for Sparse Implicit Neural Shape Completion}
\titlerunning{Observation-Conditioned Latent Energy Priors}

\author{
Paul B\"uschl\inst{1,2,6}\orcidlink{0009-0002-6685-241X}
\and
Ezequiel de la Rosa\inst{1}\orcidlink{0000-0002-9042-1962}
\and
Julia Wolleb\inst{1,3}\orcidlink{0000-0003-4087-5920}
\and
Julian McGinnis\inst{4,5}\orcidlink{0009-0000-2224-7600}
\and
C\'esar Nombela-Arrieta\inst{2}\orcidlink{0000-0003-0415-259X}
\and
Bjoern Menze\inst{1,6}\orcidlink{0000-0003-4136-5690}
}
\authorrunning{P. B\"uschl et al.}

\institute{
Department of Quantitative Biomedicine,
University of Zurich, Zurich, Switzerland\\
\email{paul.bueschl@uzh.ch}
\and
Department of Medical Oncology and Hematology,
University Hospital Zurich, Zurich, Switzerland
\and
Department of Computer Science,
ETH Zurich, Zurich, Switzerland
\and
Department of Computer Science, Technical University of Munich, Munich, Germany
\and
Munich Center for Machine Learning (MCML), Munich, Germany
\and
ETH AI Center, ETH Zurich, Zurich, Switzerland
}

\begin{document}

\maketitle

\begin{abstract}
Implicit neural representations (INRs) can model continuous 3D shapes with a shared coordinate decoder and per-instance latent codes. At test time, autodecoder-style models commonly freeze the decoder and optimize a new latent code from sparse off-grid SDF samples. When these samples underconstrain inference, the latent can drift toward regions that fit the observations but decode implausible unobserved geometry. We propose a post-hoc observation-conditioned latent energy prior for frozen INR decoders. The energy scores standardized latents conditioned on a permutation-invariant encoding of the sparse observation set and is used as a residual expert alongside an L2 latent prior selected on validation data. We evaluate on a controlled cell-nucleus SDF dataset and a public MedShapeNet-derived SDF completion dataset. The proposed L2 objective augmented with conditional energy improves consistently over a validation-selected L2 baseline in the sparsest cell-nucleus regimes and, on MedShapeNet, outperforms both L2 and a six-component GMM latent-density prior across all reported readouts. A shuffled-context ablation is consistently weaker than matched context, supporting an observation-specific contribution. These results suggest that lightweight conditional energies can make pretrained INR decoders more observation-aware without retraining.

\end{abstract}

\section{Introduction}

Implicit neural representations (INRs) model signals as continuous functions of spatial coordinates \cite{tancik2020fourier,sitzmann2020implicit,xie2022neural}, making them well suited to off-grid medical images, volumes, and shapes \cite{gropp2020implicit,molaei2023implicit,friedrich2026medfuncta,coiffier20241}. Two paradigms are commonly distinguished: INRs optimized per instance, as e.g., in MRI reconstruction \cite{shen2022nerp,huang2023neural,mcginnis2023multicontrast} and instance-wise image registration \cite{wolterink2022implicit}, and latent-conditioned INRs, which learn population-level priors \cite{xie2022neural,friedrich2026medfuncta} and thereby enable applications such as atlas construction \cite{dannecker2024cina}. In the latent-conditioned setting \cite{park2019deepsdf,dupont2022data,liu2022learning}, a decoder shared across the population receives a coordinate $\coord$ and an instance-specific latent code $\latent$ and predicts a coordinate-wise quantity such as intensity, signed distance, or a segmentation logit \cite{stoltanso2023nisf}

This decoder-plus-latent formulation underlies autodecoder shape models such as DeepSDF, which represent shapes as continuous signed-distance fields and support completion from partial or noisy observations~\cite{park2019deepsdf}. Related medical and biological work has used INRs for sparse anatomical shape reconstruction, implicit segmentation, image fitting, or cell-shape modeling~\cite{amiranashvili2022shape,stoltanso2023nisf,mcginnis2023multicontrast,wiesner2022implicit}.

In autodecoder-style INRs, training learns a shared decoder together with latent codes for the training instances. For a held-out instance, the decoder is kept fixed and inference optimizes a new latent code so that the decoded field matches the available observations. This test-time optimization is attractive for sparse shape completion: a small unordered set of off-grid SDF samples can be lifted to a full continuous field through the learned decoder. However, sparse inference is underconstrained. Many latent codes may explain the measured samples while decoding different unobserved geometry, so optimization can reduce the observation loss while moving toward a latent region that yields an implausible completion. We refer to this failure mode as \emph{latent drift}. 

The standard lightweight stabilization is an L2 penalty on the optimized latent code, which corresponds to a zero-mean isotropic Gaussian latent prior \cite{amiranashvili2022shape}. This prior is useful because it controls latent scale and gives stable tail behavior, but it is global and observation-agnostic: the same radial preference is applied regardless of which sparse points were observed. Stronger global latent-density priors, such as covariance-aware penalties, Gaussian mixtures, or learned latent energy models, can capture anisotropy, multimodality, or higher-order structure in the train-latent distribution~\cite{pang2020latent}. Yet these priors still score latent plausibility independently of the actual sparse observation set. Sparse shape completion is a conditional problem: among many plausible latent codes, inference should prefer latents that are plausible for the particular observed coordinate-value pairs. 

This leaves a gap for lightweight post-hoc priors that can be attached to frozen latent-conditioned INRs while conditioning their latent preference on sparse off-grid measurements. We therefore study whether a post-hoc conditional energy, trained from frozen-decoder latents and sparse observation sets generated by subsampling coordinate/SDF pairs, can act as an observation-aware residual prior during test-time latent optimization. The energy is trained with an Energy-Matching-inspired objective and receives a permutation-invariant encoding of the sparse observation set, following the principle that functions on unordered sets should be invariant to input order~\cite{balcerak2025energy,zaheer2017deepsets}. At inference, the decoder remains fixed and only the latent code is optimized; the conditional energy is used alongside a tuned L2 prior rather than as a replacement for it. 

We evaluate the approach in sparse SDF shape-completion settings, including a controlled cell-nucleus testbed and a public heterogeneous MedShapeNet mesh/SDF track~\cite{li2023medshapenet}. Our contributions are:
\begin{enumerate}
    \item We identify latent drift as a failure mode of sparse test-time optimization in latent-conditioned INRs.
    \item We introduce a post-hoc observation-conditioned energy prior that acts as a residual expert alongside a tuned L2 latent prior for frozen-decoder inference. 
    \item We evaluate the method against lightweight global priors and context ablations in controlled and public SDF completion settings.
\end{enumerate}
Code and configuration files are available at \url{https://github.com/pbueschl/latent-energy-priors_clean.git}.

\begin{figure}[!t]
    \centering
    \includegraphics[width=\linewidth,height=0.34\textheight,keepaspectratio]{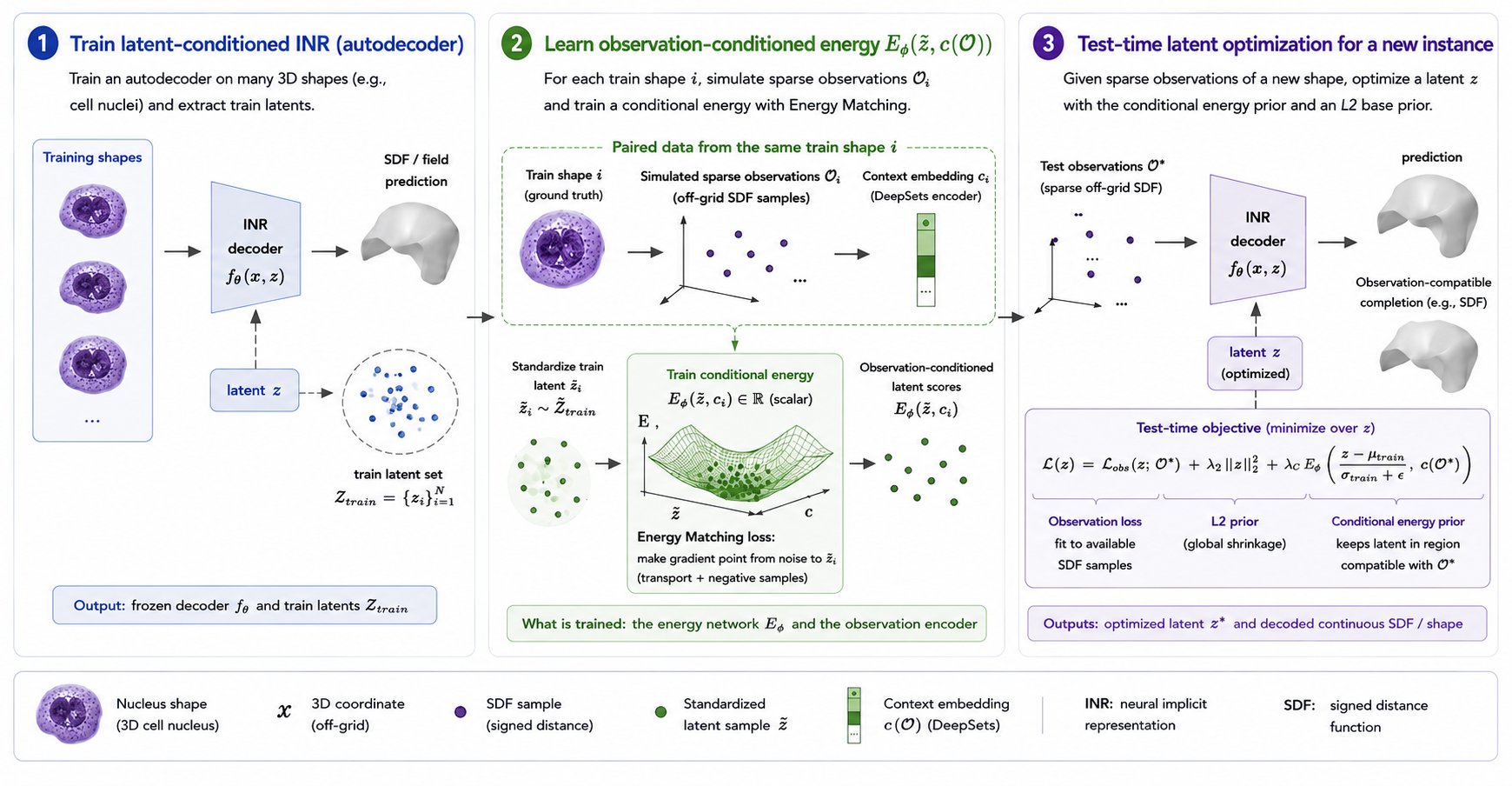}
    \caption{ Observation-conditioned latent energy prior workflow. \textit{Block~1:} a latent-conditioned INR/autodecoder provides a frozen decoder and training latents. \textit{Block~2:} sparse observation sets are generated by subsampling coordinate/SDF pairs from training shapes and paired with their latents to train a conditional scalar energy. \textit{Block~3:} for a new sparse observation set, only the latent is optimized while the decoder and energy remain fixed. The L2 prior weight is selected on validation data.}
    \label{fig:method_overview}
\end{figure}

\section{Method}

Figure~\ref{fig:method_overview} summarizes the proposed workflow. As shown in block~1, we start from a pretrained latent-conditioned INR decoder and its training latents. 
As shown in block~2, we train a post-hoc conditional energy on paired training latents and sparse observation sets generated by subsampling coordinate/SDF pairs from the same training shapes.
As shown in block~3, test-time inference keeps the decoder and energy fixed and optimizes only the latent code using the observation loss, an L2 base prior, and the conditional energy.

\subsection{Frozen Decoder and Sparse Latent Inference}

Let $f_\theta$ be a pretrained latent-conditioned decoder. In our SDF setting, $f_\theta(\coord, \latent)$ predicts the signed distance at normalized off-grid coordinate $\coord$ for an instance latent code $\latent$. 
For a new instance, we observe a sparse set $\obs^\star=\{(\coord_j,s_j)\}_{j=1}^{m}$ where $\coord_j$ is a coordinate and $s_j$ is the observed signed-distance value at that coordinate. With frozen decoder parameters, standard autodecoder inference optimizes only the latent: 
\begin{equation}
    \min_{\latent} \;
    \loss_{\mathrm{obs}}(\latent; f_\theta,\obs^\star),
    \qquad
    \loss_{\mathrm{obs}}(\latent; f_\theta,\obs^\star)
    =
    \frac{1}{m}\sum_{j=1}^{m}
    \ell_{\mathrm{SDF}}\!\left(f_\theta(\coord_j,\latent),s_j\right).
    \label{eq:obs_loss}
\end{equation}
Latent drift occurs when this optimization is underconstrained: the optimized latent fits the observed samples but decodes implausible unobserved geometry. 

\subsection{Observation-Conditioned Product-of-Experts Inference}
After training the base decoder, we extract the training latents $\mathcal{Z}_{\mathrm{train}}=\{\latent_i\}_{i=1}^{N}$ and compute per-dimension statistics $\mu_{\mathrm{train}}$ and $\sigma_{\mathrm{train}}$. 
We standardize any latent as $\stdlatent{\latent}=(\latent-\mu_{\mathrm{train}})/(\sigma_{\mathrm{train}}+\epsilon)$.
The L2 base prior is a standard squared latent penalty $R_2(\latent)$. In practice, constant normalizations of this penalty, such as division by the latent dimension, can be absorbed into the validation-selected weight $\lambda_2$.

To condition the prior on sparse observations, we encode the unordered set $\obs$ with a DeepSets-style context encoder, 
\begin{equation}
    \context_\psi(\obs)
    =
    \rho_\psi
    \left(
    \operatorname{pool}_{(\coord,s)\in\obs}
    h_\psi([\coord,s])
    \right),
    \label{eq:deepsets_context}
\end{equation}
where $h_\psi$ embeds each coordinate and SDF value, the pooling operation is invariant to the order of observations, and $\rho_\psi$ maps the pooled vector to a context vector.
The conditional energy $E_\phi:\mathbb{R}^{d}\times\mathbb{R}^{d_c}\rightarrow\mathbb{R}$ is a scalar network that scores a standardized latent in the context of this observation embedding, where $d$ is the latent dimensionality and $d_c$ is the context dimension.

The main test-time objective is 
\begin{equation}
    \min_{\latent}\;
    \loss_{\mathrm{obs}}(\latent; f_\theta,\obs^\star)
    +
    \lambda_2 R_2(\latent)
    +
    \lambda_C
    E_\phi\!\left(
        \stdlatent{\latent},
        \context_\psi(\obs^\star)
    \right).
    \label{eq:energy_inference}
\end{equation}
The L2 expert controls latent scale, while the conditional energy acts as an observation-aware residual expert. The decoder parameters $\theta$ are frozen; only the latent code $\latent$ is optimized during inference. 

\subsection{Training the Conditional Energy}
The conditional energy and context encoder are trained after the decoder has been fitted. For each training shape $i$, we pair its training latent $\latent_i$ with sparse observation sets $\obs_i$ generated by subsampling coordinate/SDF pairs from the same training shape. Each minibatch samples the number of observed SDF samples $m$ from the dataset-specific training values and then subsamples $m$ coordinate/SDF pairs. The conditional energy is trained only from train-split latents and train-split observations. 

We adapt Energy Matching to the conditional latent space. Let $\context_i=\context_\psi(\obs_i)$ denote the observation context and $\tilde{\latent}_i=\stdlatent{\latent_i}$ the standardized training latent.
We sample Gaussian noise $\tilde z_0\sim\mathcal{N}(0,I)$ and $t\sim\mathcal{U}(0,1)$, form the interpolant $\tilde{\latent}_t=(1-t)\tilde{\latent}_0+t\tilde{\latent}_i$, and regress the energy gradient so that the negative gradient points toward the matched training latent:
\begin{equation}
    \mathcal{L}_{\mathrm{CEM}}
    =
    \mathbb{E}
    \left[
    \frac{1}{d}
    \left\|
    \nabla_{\tilde z_t}
    E_\phi(\tilde z_t,c_i)
    +
    (\tilde z_i-\tilde z_0)
    \right\|_2^2
    \right].
    \label{eq:conditional_energy_matching}
\end{equation}

We add ranking losses to make the matched latent-context pair lower energy than mismatched pairs. For $k\neq i$ and margin $m_{\mathrm{rank}}$,
\begin{equation}
\begin{aligned}
    \mathcal{L}_{\mathrm{rank}}
    =
    &\;
    \mathbb{E}_{i,k}
    \left[
    \operatorname{softplus}
    \left(
    E_\phi(\tilde z_i,c_i)
    +m_{\mathrm{rank}}
    -E_\phi(\tilde z_k,c_i)
    \right)
    \right]
    \\
    &+
    \mathbb{E}_{i,k}
    \left[
    \operatorname{softplus}
    \left(
    E_\phi(\tilde z_i,c_i)
    +m_{\mathrm{rank}}
    -E_\phi(\tilde z_i,c_k)
    \right)
    \right].
\end{aligned}
\label{eq:rank_loss}
\end{equation}
The first term compares the correct latent with a mismatched latent under the same context; the second compares the correct context with a mismatched context for the same latent. Together, they discourage the energy from ignoring either the latent or the observation context. Finally, we include off-manifold negatives $\tilde z^{-}$, initialized from
broad latent noise or train-latent perturbations and refined by short Langevin
chains. These negatives are used only during energy training. Let
\begin{equation}
    \bar E_{\mathrm{pos}}
    =
    \frac{1}{B}
    \sum_{i=1}^{B}
    E_\phi(\tilde z_i,c_i)
\end{equation}
denote the average matched-pair energy in the current minibatch. The negative
calibration loss is
\begin{equation}
    \mathcal{L}_{\mathrm{neg}}
    =
    \mathbb{E}
    \left[
    \operatorname{softplus}
    \left(
    \bar E_{\mathrm{pos}}
    +m_{\mathrm{neg}}
    -E_\phi(\tilde z^{-},c_i)
    \right)
    \right].
    \label{eq:negative_loss}
\end{equation}
Thus, valid off-manifold negatives are encouraged to have higher energy than the average matched train pair in the current minibatch. 

The full training loss is 
\begin{equation}
    \mathcal{L}_{\mathrm{energy}}
    =
    \alpha_{\mathrm{CEM}}\mathcal{L}_{\mathrm{CEM}}
    +
    \alpha_{\mathrm{rank}}\mathcal{L}_{\mathrm{rank}}
    +
    \alpha_{\mathrm{neg}}\mathcal{L}_{\mathrm{neg}}
    +
    \alpha_{\mathrm{scale}}
    \mathbb{E}_{(\tilde z,c)\in\mathcal{B}_{\mathrm{energy}}}
    \left[
    E_\phi(\tilde z,c)^2
    \right].
    \label{eq:energy_training_loss}
\end{equation}

Here, $\mathcal{B}_{\mathrm{energy}}$ denotes the set of energy evaluations used in the current training step, including matched pairs, mismatched latent-context pairs, and negative samples. The interpolated latents and Langevin-refined negatives are training-only samples; test-time inference uses the learned energy as a differentiable prior.

\section{Experiments}
\subsection{Datasets, Sparse Observations, and Metrics}
We evaluate sparse SDF completion in a controlled cell-nucleus setting and a public MedShapeNet-derived SDF setting. The number of observed off-grid SDF samples is denoted by $m$; all remaining samples are held out for SDF-error evaluation. The nucleus split contains 2,924/699/492 train/validation/locked-test nuclei; each training nucleus contributes one latent code. Each nucleus has 4,096 off-grid SDF samples, and sparse inference uses $m\in\{16,32,64,128\}$.

MedShapeNet is a public collection of anatomical 3D meshes. We use bladder, brain, heart, liver, skull MRI, and vertebrae with a balanced 960/120/120 train/validation/test split. Meshes are normalized and converted to clipped off-grid SDF samples for autodecoder training and sparse inference with $m\in\{16,32,64,128,256\}$.

The primary metric is held-out SDF error on unobserved off-grid samples. Secondary grid readouts include Grid SDF MAE and occupancy Dice/IoU after thresholding the SDF at zero; the nucleus setting also reports surface Dice, which measures boundary agreement within the fixed distance tolerance of 0.1 in normalized coordinate units.

\subsection{Baselines, Ablations, and Inference Protocol}
All methods use the same frozen decoder, matched sparse observations, latent initialization, optimization budget, and evaluation points within each setting. We distinguish baselines from ablations. The baselines are: No prior, which minimizes only $\mathcal{L}_{\mathrm{obs}}$; Tuned L2, which adds the validation-selected squared latent penalty; and GMM, a global latent-density baseline that fits a six-component Gaussian mixture to standardized training latents and uses $-\log p_{\mathrm{GMM}}(\tilde z)$ as a prior ($\lambda_{\mathrm{GMM}}=10^{-5}$). Our main method is L2 + Conditional Energy. The ablations are Conditional Energy only, which removes the L2 expert, and shuffled context, which evaluates the conditional energy with a mismatched observation context. For nuclei, latent optimization runs for 1,000 Adam steps with learning rate $10^{-2}$ and latent dimension $d=64$. Grid-based metrics use a $32^3$ grid over normalized coordinates $[-1.5,1.5]^3$. For MedShapeNet, locked-test inference runs for 2,000 Adam steps. Prior weights are selected by validation grid search using the setting-specific selection metric, and the locked test is evaluated once after freezing the selected weights and hyperparameters. For nuclei, final-step surface Dice selects $\lambda_2=0.3$, $\lambda_C=3{\times}10^{-6}$, and $\lambda_2+\lambda_C=0.3+10^{-5}$. For MedShapeNet, validation Eval L1 selects $\lambda_2=10^{-3}$, $\lambda_C=10^{-5}$, and $\lambda_2+\lambda_C=10^{-3}+3{\times}10^{-5}$. We report paired bootstrap confidence intervals; for nuclei, we additionally use paired Wilcoxon tests with FDR correction.
\FloatBarrier
\section{Results}
\subsection{Locked-Test Performance}
On the locked nucleus test set, L2 + Conditional Energy gives the best or tied-best performance across sparse observation settings (Table~\ref{tab:nucleus_results}). Conditional-Energy-only is reported as an ablation: it tests whether the learned energy can replace the L2 expert. It improves over No prior in sparse settings but remains below Tuned L2, supporting the product-of-experts design in which the conditional energy complements rather than replaces the L2 prior.
\begin{table}[!t]
\centering
\caption{Nucleus locked-test performance. CondE: Conditional Energy; Ours: L2 plus CondE.}
\label{tab:nucleus_results}
\fontsize{8}{9}\selectfont
\setlength{\tabcolsep}{2.5pt}
\begin{tabular}{@{}c cccc cccc@{}}
\toprule
& \multicolumn{4}{c}{Grid SDF MAE $\downarrow$}
& \multicolumn{4}{c}{Surface Dice $\uparrow$} \\
\cmidrule(lr){2-5}\cmidrule(lr){6-9}
Obs. & No prior & CondE & Tuned L2 & Ours
     & No prior & CondE & Tuned L2 & Ours \\
\midrule
16  & 0.1026 & 0.0968 & 0.0851 & \textbf{0.0838}
    & 0.7818 & 0.8143 & 0.8590 & \textbf{0.8650} \\
32  & 0.0756 & 0.0724 & 0.0685 & \textbf{0.0664}
    & 0.8804 & 0.8986 & 0.9138 & \textbf{0.9222} \\
64  & 0.0606 & 0.0581 & 0.0558 & \textbf{0.0551}
    & 0.9336 & 0.9412 & 0.9521 & \textbf{0.9545} \\
128 & 0.0501 & 0.0496 & \textbf{0.0485} & \textbf{0.0485}
    & 0.9667 & 0.9677 & 0.9725 & \textbf{0.9728} \\
\bottomrule
\end{tabular}
\end{table}

On MedShapeNet, the global GMM prior often improves over Tuned L2, indicating that modeling train-latent density is useful. However, L2 + Conditional Energy gives the best performance for all reported MedShapeNet readouts and numbers of observed samples (Table~\ref{tab:med_results}). This suggests that observation conditioning adds information beyond a global multimodal latent prior.
\begin{table}[!t]
\centering
\caption{MedShapeNet locked-test performance. GMM K6: six-component global latent mixture; Ours: L2+CondE.}
\label{tab:med_results}
\fontsize{8}{9}\selectfont
\setlength{\tabcolsep}{4pt}
\begin{tabular}{@{}c l cccc@{}}
\toprule
Obs. & Method & Eval L1 $\downarrow$ & Grid SDF MAE $\downarrow$ & Dice $\uparrow$ & IoU $\uparrow$ \\
\midrule
16  & Tuned L2 & 0.0252 & 0.1156 & 0.5102 & 0.3666 \\
    & GMM K6   & 0.0246 & 0.1125 & 0.5268 & 0.3855 \\
    & Ours     & \textbf{0.0245} & \textbf{0.1079} & \textbf{0.5500} & \textbf{0.4056} \\
\midrule
32  & Tuned L2 & 0.0212 & 0.1046 & 0.5854 & 0.4425 \\
    & GMM K6   & 0.0207 & 0.0998 & 0.6019 & 0.4587 \\
    & Ours     & \textbf{0.0200} & \textbf{0.0919} & \textbf{0.6271} & \textbf{0.4863} \\
\midrule
64  & Tuned L2 & 0.0172 & 0.0917 & 0.6668 & 0.5304 \\
    & GMM K6   & 0.0165 & 0.0867 & 0.6847 & 0.5509 \\
    & Ours     & \textbf{0.0163} & \textbf{0.0814} & \textbf{0.6908} & \textbf{0.5571} \\
\midrule
128 & Tuned L2 & 0.0129 & 0.0754 & 0.7495 & 0.6253 \\
    & GMM K6   & 0.0129 & 0.0759 & 0.7505 & 0.6276 \\
    & Ours     & \textbf{0.0125} & \textbf{0.0697} & \textbf{0.7632} & \textbf{0.6424} \\
\midrule
256 & Tuned L2 & 0.0096 & 0.0624 & 0.8156 & 0.7089 \\
    & GMM K6   & 0.0094 & 0.0614 & 0.8208 & 0.7156 \\
    & Ours     & \textbf{0.0093} & \textbf{0.0571} & \textbf{0.8237} & \textbf{0.7199} \\
\bottomrule
\end{tabular}
\end{table}
\subsection{Paired Gains over Global Priors}
\begin{figure}[!t]
    \centering
    \includegraphics[width=\linewidth]{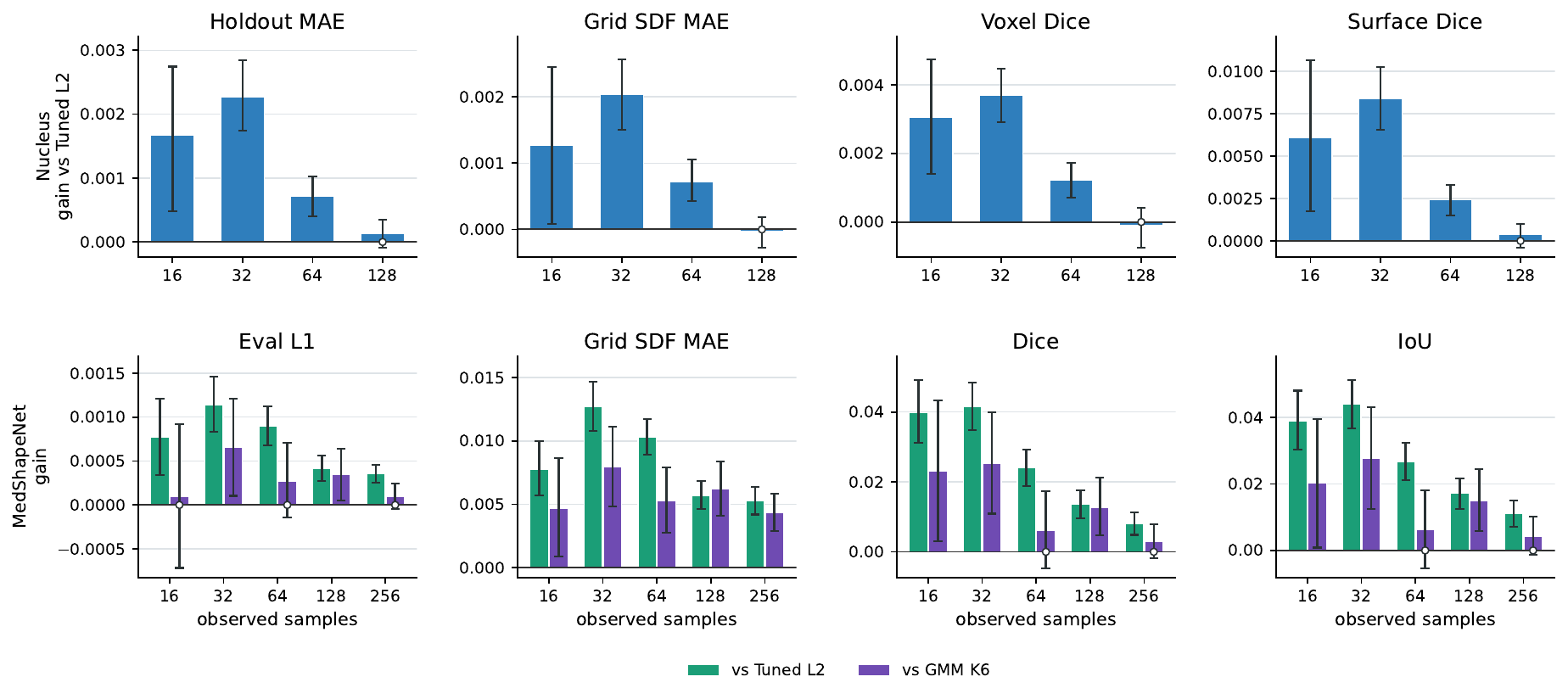}
    \caption{Paired locked-test gains of L2 + Conditional Energy over Tuned L2 and, for MedShapeNet, over GMM K6. Positive values denote error reduction or score increase. Error bars show 95\% paired-bootstrap confidence intervals; open markers cross zero.}
    \label{fig:combined_l2_conditional_vs_l2_ci}
\end{figure}

Figure~\ref{fig:combined_l2_conditional_vs_l2_ci} reports paired locked-test gains. Positive values denote error reduction for L1/MAE metrics and score increase for Dice/IoU metrics. For nuclei, L2 + Conditional Energy improves most clearly over Tuned L2 at 16 and 32 observed samples, and gains shrink as observations become denser. FDR-corrected paired Wilcoxon tests show significant improvements over Tuned L2 for all four nucleus metrics at 16, 32, and 64 samples, with largest corrected $p$-value $1.38{\times}10^{-2}$; at 128 samples, only surface Dice remains significant ($p=8.0{\times}10^{-4}$). On MedShapeNet, L2 + Conditional Energy improves over Tuned L2 across 16--256 samples and improves over GMM K6 most clearly for Grid SDF MAE and overlap-based readouts.
\subsection{Context Ablation and Qualitative Examples}
\begin{figure}[t]
    \centering
    \includegraphics[width=0.95\linewidth]{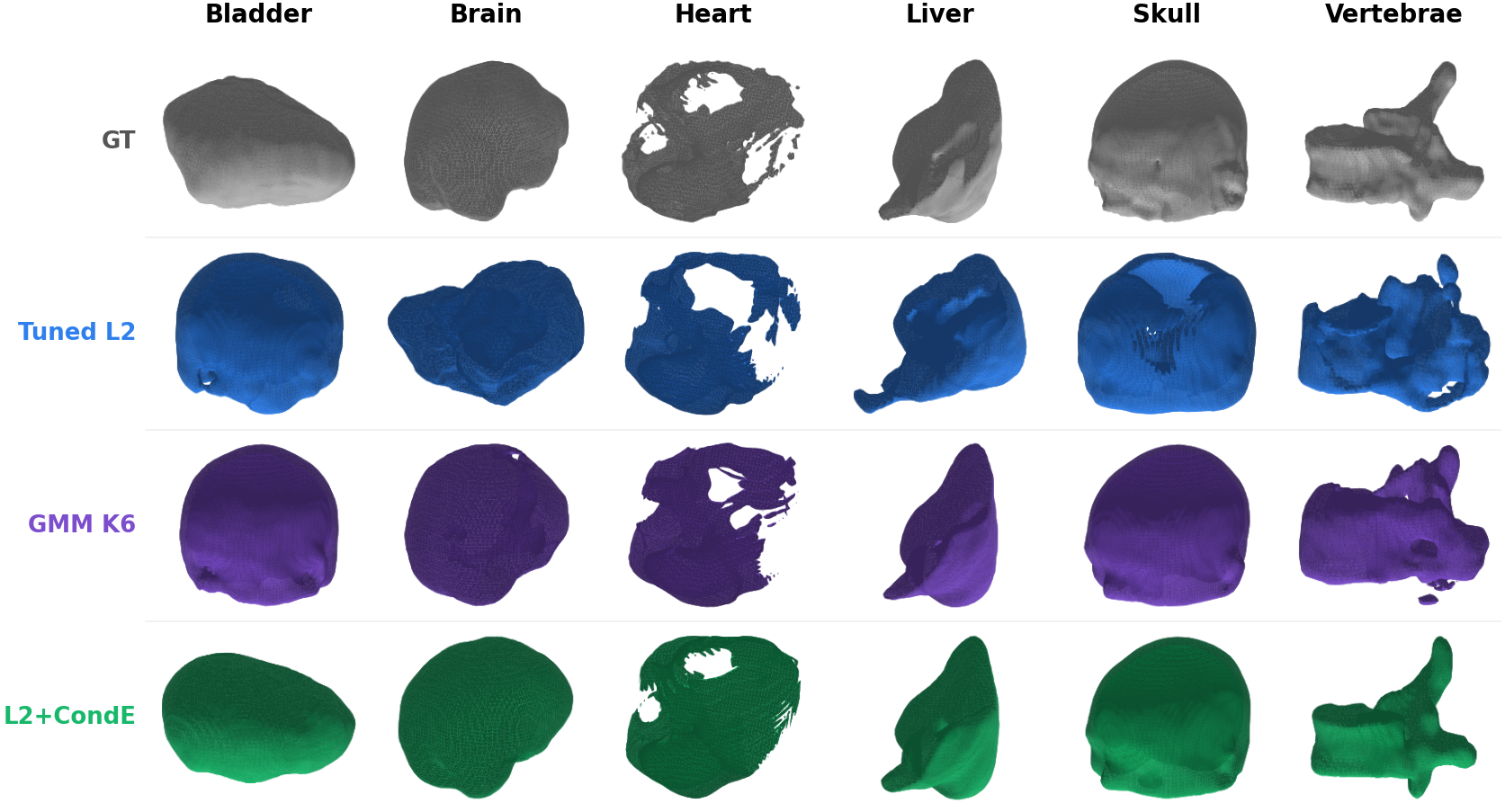}
    \caption{Qualitative MedShapeNet reconstructions from sparse observations. Rows show ground truth, Tuned L2, GMM K6, and L2 + Conditional Energy. Columns show anatomical categories.}
    \label{fig:medshape_qualitative}
\end{figure}

The shuffled-context ablation tests whether the learned energy uses the matched observation set or only acts as an additional global prior. On MedShapeNet, shuffled context remains better than Tuned L2 on Grid SDF MAE, indicating a general latent-prior component. However, matched context improves over shuffled context at every observed-sample count, reducing Grid SDF MAE by 0.0026--0.0072 relative to shuffled context. This supports an observation-specific contribution of the DeepSets context. Figure~\ref{fig:medshape_qualitative} shows representative MedShapeNet reconstructions. Compared with Tuned L2 and GMM K6, L2 + Conditional Energy more consistently preserves the category-compatible shape structure under the same sparse observations.
\begin{table}[!t]
\centering
\caption{MedShapeNet shuffled-context ablation on Grid SDF MAE. Lower is better. $\Delta$ values denote reductions relative to Tuned L2.}
\label{tab:shuffled_context}
\fontsize{8}{9}\selectfont
\setlength{\tabcolsep}{5pt}
\begin{tabular}{@{}c ccccc@{}}
\toprule
Obs. & Tuned L2 & Ours & Ours shuf. & $\Delta$ matched & $\Delta$ shuf. \\
\midrule
16  & 0.1156 & \textbf{0.1079} & 0.1124 & 0.0077 & 0.0032 \\
32  & 0.1046 & \textbf{0.0919} & 0.0991 & 0.0127 & 0.0055 \\
64  & 0.0917 & \textbf{0.0814} & 0.0868 & 0.0103 & 0.0049 \\
128 & 0.0754 & \textbf{0.0697} & 0.0723 & 0.0057 & 0.0031 \\
256 & 0.0624 & \textbf{0.0571} & 0.0600 & 0.0053 & 0.0023 \\
\bottomrule
\end{tabular}
\end{table}

\section{Discussion \& Conclusion}
The main takeaway is that observation conditioning is most useful when sparse test-time inference is underconstrained. Tuned L2 provides stable global shrinkage, while the conditional Energy acts as an observation conditioned complement. The shuffled context preserves a general latent-prior effect, while the matched context improves on it at all tested observation counts, which supports the additional observation-specific contribution interpretation. The GMM baseline indicates that modeling global train-latent density is useful on heterogeneous MedShapeNet. 

\begin{figure}[!t]
    \centering
    \includegraphics[width=\linewidth]{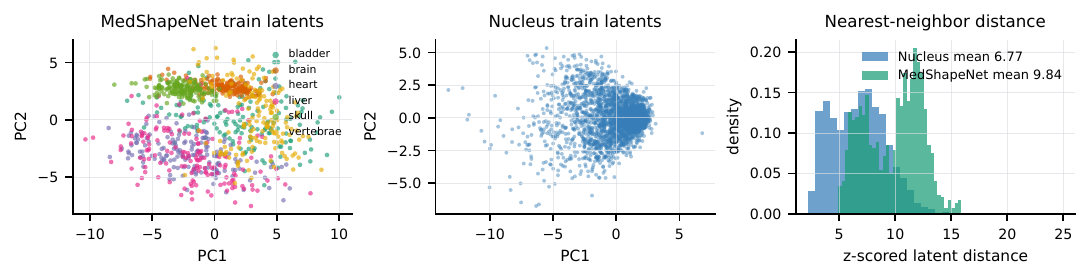}
    \caption{Standardized latent-space comparison. MedShapeNet latents show stronger category structure and larger nearest-neighbor distances than nucleus latents. This is consistent with, but does not establish, the hypothesis that a conditional energy is useful when one isotropic L2 prior is too coarse.}
    \label{fig:latent_space_heterogeneity}
\end{figure}

The latent space-comparison in Fig.~\ref{fig:latent_space_heterogeneity} provides a descriptive view of the two latent distributions. Looking at the per-dimension standardized latents, MedShapeNet shows stronger category structure and larger nearest-neighbor distance than the nucleus dataset. This is consistent with the results that show that a single isotropic L2 prior is too coarse, especially for the latent space of the more heterogeneous MedShapeNet dataset. Together with the improvement of L2+CondE over GMM K6 on the reported MedShapeNet readouts, this suggests that the conditional energy adds information beyond a global multimodal latent prior.
Several limitations remain. We evaluate one controlled cell-nucleus SDF setting and one public MedShapeNet-derived SDF setting. Broader context randomization, out-of-distribution tests, and comparisons to additional structured priors beyond GMM, such as PCA-Gaussian, normalizing-flow, or diffusion priors, remain important future work. Adaptive prior weights or uncertainty estimates may also help detect cases where sparse observations conflict with the learned prior.

These results suggest that lightweight post-hoc priors can make pretrained INR decoders more observation-aware without retraining. Future work will study stronger conditional energy architectures, realistic observation patterns like partial surfaces or samples and broader medical and non-medical INR tasks, including topology-rich reconstruction problems such as vessel trees.

\FloatBarrier

\begin{credits}
\subsubsection{\ackname}
This work has been supported by the Helmut Horten Foundation.

\subsubsection{\discintname}
The authors have no competing interests to declare that are relevant to the
content of this article.
\end{credits}
\clearpage

\bibliographystyle{splncs04}
\bibliography{refs}
\end{document}